\documentclass{article}

  \usepackage[dblblindworkshop, final]{neurips_2025}
\workshoptitle{AI4Mat}

\usepackage[utf8]{inputenc} 
\usepackage[T1]{fontenc}    
\usepackage{amsfonts}      

\usepackage{graphicx}      
\usepackage{xcolor}        
\usepackage{url}           
\usepackage{microtype}     
\usepackage{nicefrac}     

\usepackage{booktabs}       
\usepackage{multirow}       

\usepackage{natbib}         % Стили библиографии. Важно загрузить ДО hyperref
\usepackage{hyperref} 
\title{Benchmarking Agentic Systems in Automated Scientific Information Extraction with ChemX}
\workshoptitle{AI4Mat}

\author{%
  Anastasia Vepreva$^{1}$ \quad  Julia Razlivina$^{1}$ \smallskip \\ 
\textbf{Maria Eremeeva$^{1}$ \quad Nina Gubina$^{1}$ \quad Anastasia Orlova$^{1}$ \quad Aleksei Dmitrenko$^{1}$}  \smallskip \\ 
\textbf{Ksenya Kapranova$^{1}$ \quad Susan Jyakhwo$^{1}$ \quad Nikita Vasilev$^{1}$ \quad Arsen Sarkisyan$^{1}$}  \smallskip \\ 
\textbf{Ivan Yu. Chernyshov$^{1}$ \quad Vladimir Vinogradov$^{1}$ \quad Andrei Dmitrenko$^{1,2}$}  \medskip \\
$^{1}$Center for AI in Chemistry, ITMO University, St. Petersburg, Russia  \smallskip \\
$^{2}$D ONE AG, Zurich, Switzerland  \smallskip \\
 \smallskip \\
\texttt{dmitrenko@pish.itmo.ru}}

\begin{document}

\maketitle

\begin{abstract}
  The emergence of agent-based systems represents a significant advancement in artificial intelligence, with growing applications in automated data extraction. However, chemical information extraction remains a formidable challenge due to the inherent heterogeneity of chemical data. Current agent-based approaches, both general-purpose and domain-specific, exhibit limited performance in this domain. To address this gap, we present ChemX, a comprehensive collection of 10 manually curated and domain-expert-validated datasets focusing on nanomaterials and small molecules. These datasets are designed to rigorously evaluate and enhance automated extraction methodologies in chemistry. To demonstrate their utility, we conduct an extensive benchmarking study comparing existing state-of-the-art agentic systems such as ChatGPT Agent and chemical-specific data extraction agents. Additionally, we introduce our own single-agent approach that enables precise control over document preprocessing prior to extraction. We further evaluate the performance of modern baselines, such as GPT-5 and GPT-5 Thinking, to compare their capabilities with agentic approaches. Our empirical findings reveal persistent challenges in chemical information extraction, particularly in processing domain-specific terminology, complex tabular and schematic representations, and context-dependent ambiguities. The ChemX benchmark serves as a critical resource for advancing automated information extraction in chemistry, challenging the generalization capabilities of existing methods, and providing valuable insights into effective evaluation strategies.
\end{abstract}

\section{Introduction}

Over the past decade, machine learning has significantly advanced chemical discovery, underscoring the need for well-structured data \citep{Butler2018, Moosavi2020, SanchezLengeling2018}. Standardized datasets provide essential metrics for comparing algorithms, identifying their limitations, and accelerating progress \citep{Gaulton2011, Groom2016, Burley2017, Wu2018, Ganscha2025}.  However, major gaps persist, particularly in specialized domains, creating an urgent need for robust systems to automatically extract and curate chemical data from diverse sources. 

While conventional NLP methods have been used for named entity recognition in the sciences \citep{https://doi.org/10.48550/arxiv.2010.09885, Yu2022, Wang2019}, they remain limited in the broader range of tasks required for a chemical data extraction tool. Recent advances in large language models (LLMs) have demonstrated remarkable improvements in contextual understanding and reasoning \citep{https://doi.org/10.48550/arxiv.2308.08155}. Autonomous multi-agent systems are becoming a new frontier in the automation of scientific research \citep{https://doi.org/10.48550/arxiv.2411.04468, Lu2024}. Recent advances in automated chemical information extraction have increasingly leveraged agentic AI approaches, which employ autonomous, goal-directed agents capable of reasoning, planning, and executing complex workflows \citep{MBran2024, Boiko2023, https://doi.org/10.48550/arxiv.2312.07559}. These agentic systems differ fundamentally from traditional AI methods by integrating domain-specific knowledge with capabilities for contextual understanding and iterative decision-making.  Currently, highly specialized systems exist for data extraction in materials science, as well as for the extraction of organic reaction data or deep eutectic solvent knowledge \citep{Li2025, Ansari2024, https://doi.org/10.48550/arxiv.2401.17244, Mullick2024, Odobesku2025, https://doi.org/10.48550/arxiv.2404.01462, Peng2025}.  Applying multi-agent systems to chemical data extraction remains challenging due to domain adaptation, making it an essential research challenge. To support it, we present ChemX, a manually curated multimodal benchmark dataset aimed at extracting chemical features from textual and visual content across diverse chemical domains. By capturing the heterogeneity and interconnectedness of real-world chemical literature, ChemX provides a foundation for evaluating automation extraction systems. This work makes two major contributions:

\begin{itemize}
    \item We provide the ChemX benchmark, a collection of 10 curated datasets describing various properties of nanomaterials and small molecules. Each dataset is accompanied with detailed documentation, standardized metadata, and cross-verification by domain experts. The datasets are hosted as a collection on the Hugging Face. The accompanying documentation will be provided separately to ensure compliance with anonymization guidelines. 
    
    \item We present a systematic evaluation of state-of-the-art agentic systems in the task of automated information extraction from domain-specific scientific literature. The code for the extraction experiments is provided in the \href{https://ai-chem.github.io/ChemX/index.html}{https://ai-chem.github.io/ChemX}.
\end{itemize}

\section{Related works}

Recent years have seen a growing ecosystem of chemical science benchmarks, many focusing on machine learning for property prediction, structural analysis, or vision-language tasks \citep{https://doi.org/10.48550/arxiv.2406.14347, NEURIPS2022_ada36dfe, NEURIPS2024_e38e60b3}. However, these are not designed for evaluating automated information extraction systems. The closest related study, nanoMINER \citep{Odobesku2025}, demonstrates structured extraction but is limited to one dataset related to nanozymes. We address this gap with 10 diverse datasets, benchmarking modern LLMs and agentic systems, including nanoMINER for comparison.

\section{ChemX}

\begin{figure}[h!]
    \centering
    \includegraphics[width=0.9\linewidth]{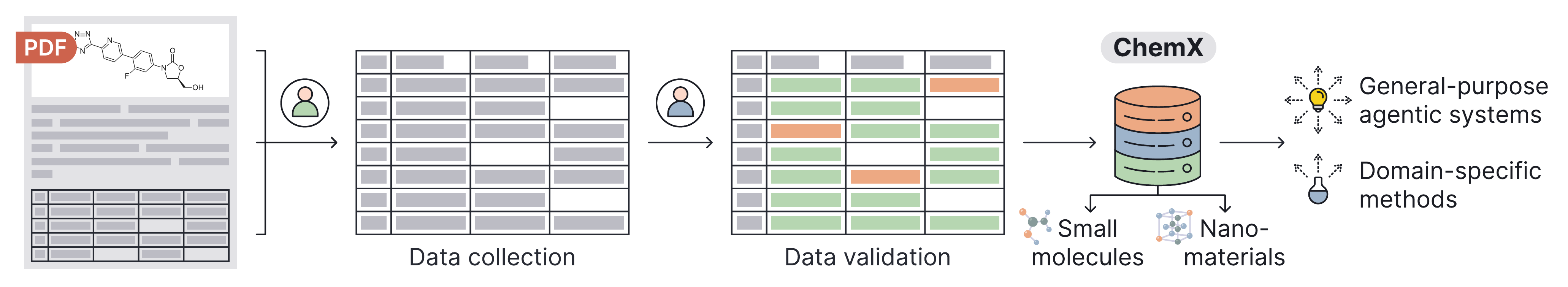}
    \caption{ChemX. This pipeline includes manual collection of multimodal data from scientific articles, further validation by domain experts and benchmarking automated data extraction.}
    \label{fig:main}
\end{figure}

ChemX is a comprehensive multimodal benchmark comprising 10 rigorously validated datasets spanning two major chemical domains: nanomaterials and small molecules (\autoref{fig:main}). The collection is designed to support robust automated information extraction across heterogeneous data types, including tables, graphs, and unstructured text. 

The datasets' ontology varies between domains:

\begin{itemize}
    \item Small molecule datasets focus on molecular descriptors like SMILES representations, biological activity metrics (MIC, IC50), and compound metadata
    \item Nanomaterial datasets encompass a broader range of parameters, including physicochemical properties, synthesis conditions, structural characteristics, and application-specific outcomes
\end{itemize}

The more detailed description of each dataset,  quality control process and dataset analysis are presented in Appendix (\autoref{ontology}, \autoref{Quality Control}, \autoref{Datasets Overview and Analysis}).  Including the datasets of varying sizes and complexity in both domains creates a balanced and practical benchmark for automated information extraction. 

All datasets were labeled by complexity level, which is described in detail in the \autoref{Selection of articles for analysis}. 

\section{Experiments}

\subsection{Information extraction task}

This study was designed to evaluate modern agentic information extraction approaches using datasets from ChemX. We selected two datasets of the lowest complexity within the domain, as categorized in \autoref{table_atricles}, namely, nanozymes (nanomaterials) and chelate complexes (small molecules). \autoref{Datasets Overview and Analysis} demonstrates that closed-access articles constitute the vast majority within each dataset. To ensure the selection was both representative and reproducible, we included two open-access articles for analysis (\autoref{selected_articles}).  An end-to-end information extraction task is, therefore, defined as follows: given the article file (or DOI, in case an attachment is not supported), output the extracted information according to the instructions in the prompt.

\subsection{Methods and metrics}

A detailed description of the prompts and metrics used to evaluate the quality of extraction is described in the \autoref{metrics}. The latest models GPT-5 and GPT-5 Thinking were selected as baselines. Agent-based approaches were also implemented, encompassing both a general-purpose ChatGPT Agent and domain-specific systems optimized for data extraction in singular and multiple domains such as FutureHouse \citep{futurehouseFutureHouse}, SLM-Matrix \citep{Li2025}, Eunomia Agent \citep{Ansari2024}, ChemOpenIE \citep{https://doi.org/10.48550/arxiv.2404.01462}, and nanoMINER systems. 

\subsection{Single-agent approach}

To address OpenAI's opaque PDF/screenshot processing, which risks inconsistent extractions, we develop a single-agent approach for structured text conversion, ensuring reproducibility and semantic integrity. Using marker-pdf SDK \citep{githubGitHubVikParuchurimarker}, we extract text blocks, tables, and images, preserving document structure. Text and tables are converted to markdown, while images are replaced with local paths. Extracted images are processed by GPT-4o (2024-11-20) to generate descriptions, inserted into markdown at original locations via \texttt{<DESCRIPTION\_FROM\_IMAGE>} tags. The final markdown file is then processed by GPT-4.1, GPT-5, and GPT-OSS-20b for extraction, with results consolidated into dataset-specific CSV files. 

\section{Results and Discussion}

As presented in \autoref{table_metrics}, which details the average extraction metrics across all dataset columns, the general methods demonstrated superior performance for both nanomaterial and small molecule datasets. A notable exception is the nanoMINER method, which achieved the highest metrics; however, its applicability is severely limited by its specificity to a single dataset. Contrary to expectations, the GPT-5 Thinking model configured for extended reasoning demonstrates inferior performance on the extraction task compared to standard GPT-5. 

\begin{table}[ht!]
\centering
\caption{Extraction metrics. * ChatGPT Agent fails to complete the extraction task for the nanozymes dataset due to alleged policy violations. ** NanoMINER was originally designed to work with the nanozymes dataset only and cannot generalize.}
\label{table_metrics}
\renewcommand{\arraystretch}{1.2}
\begin{tabular}{lcccccc}
\hline
\multicolumn{1}{c}{\multirow{2}{*}{Method}} & \multicolumn{3}{c}{Nanozymes} & \multicolumn{3}{c}{Complexes} \\ \cline{2-7} 
\multicolumn{1}{c}{}   & Precision     & Recall        & F1            & Precision     & Recall        & F1            \\ \hline
GPT-5                  & 0.33          & 0.53          & 0.37          & 0.45          & 0.18          & 0.23          \\
GPT-5 Thinking         & 0.01          & 0.04          & 0.02          & 0.22          & 0.18          & 0.19          \\ \hline
Single-agent (GPT-4.1) & 0.41          & 0.73          & 0.52          & 0.35          & 0.21          & 0.27          \\
Single-agent (GPT-5)   & 0.47          & \textbf{0.75} & 0.58          & 0.32          & 0.39          & 0.35          \\
Single-agent (GPT-OSS) & 0.56          & 0.67          & 0.61 & 0.36          & 0.31          & 0.33          \\
ChatGPT Agent*              & -             & -             & -             & \textbf{0.50} & \textbf{0.42} & \textbf{0.46} \\
SLM-Matrix             & 0.14          & 0.55          & 0.22          & 0.40          & 0.38          & 0.39          \\
FutureHouse            & 0.05          & 0.31          & 0.09          & 0.12          & 0.06          & 0.06          \\
NanoMINER**              & \textbf{0.90} & 0.74          & \textbf{0.80}          & -             & -             & -             \\ \hline
\end{tabular}
\end{table}

Other domain-specific multi-agent systems, such as SLM Matrix (designed for material data extraction using small language models) and FutureHouse, were found to be inadequate for the specified extraction task. Among the general methods, a consistent pattern emerged: performance was stronger on nanomaterial data. This is despite the inherent complexity of these datasets, as detailed in \autoref{Selection of articles for analysis}, \autoref{table_atricles}. This superior performance is likely attributable to a critical shortcoming in small molecule extraction: the inability of all evaluated systems to accurately extract SMILES notations, as they lack integrated tools for converting molecular images to SMILES strings. Consequently, the reported metrics for small molecules may be systematically underestimated; complete per-column metrics for both datasets are provided in \autoref{Results and Discussion}. 

Further analysis reveals that the single-agent approach yielded better results than the baseline models. Notably, pre-processing documents into structured text significantly enhanced extraction quality, improving recall from 0.53 to 0.75 for the GPT-5 model. In contrast, ChatGPT Agent issued warnings concerning terms of use violations when processing the nanozyme dataset, though it did achieve the best metrics on the small molecule dataset.

\begin{table}[ht!]
\centering
\caption{Agentic extraction systems overview.}
\label{table_systems}
\renewcommand{\arraystretch}{1.2}
\begin{tabular}{lccccc}
\hline
\multicolumn{1}{c}{Method} &
  PDF file &
  \begin{tabular}[c]{@{}c@{}}Output format\end{tabular} &
  Generalizability &
  \begin{tabular}[c]{@{}c@{}}End-to-end\end{tabular} &
  Multimodality \\ \hline
Single-agent (ours) & yes         & yes         & yes         & yes         & yes \\
ChatGPT Agent & yes        & \textbf{no} & \textbf{no}   & yes  & yes \\
SLM-Matrix  & yes         & yes         & yes         & yes         & yes \\
NanoMINER   & yes         & yes         & \textbf{no} & yes         & yes \\
FutureHouse & \textbf{no} & yes         & yes         & yes         & yes \\
Eunomia     & yes         & \textbf{no} & \textbf{no} & \textbf{no} & yes \\
OpenChemIE  & yes         & yes         & \textbf{no} & \textbf{no} & yes \\ \hline
\end{tabular}
\end{table}

The pronounced methodological disparities among specialized approaches introduced additional complexities. We qualitatively evaluated these methods based on key properties, including the ability to process PDF files, adherence to a specified output structure, multimodality, generalizability, and the capacity to complete a full extraction task (i.e., extracting all required data fields). Approaches incapable of executing the full (end-to-end) extraction task were excluded from the analysis (\autoref{table_systems}). For instance, the OpenChemIE method, designed for extracting organic chemical reactions, was omitted as it only extracts molecular identifiers (ID) and SMILES notations. Similarly, the Eunomia method, developed for materials science data extraction, was excluded due to its failure to produce a correct output file structure.

Our findings demonstrate that, despite recent advances in AI and agentic systems, the accurate extraction of chemical information remains a surprisingly complex task that requires much of innovation to be solved. As automated information extraction increasingly relies on multi-agent frameworks, greater research emphasis should be placed on agent orchestration. As the first resource of its kind, ChemX provides a foundation for advancing automated information extraction in chemistry, enabling systematic evaluation and refinement of new techniques.

\section{Conclusion}

ChemX is an expert-curated, multimodal benchmark for chemical information extraction, addressing gaps in existing resources through standardized schemas, domain diversity, and provenance metadata. Its utility was demonstrated by evaluating state-of-the-art agentic systems compared against the leading reasoning LLMs. As the first benchmarking resource of its kind, ChemX provides a critical foundation for advancing automated information extraction in chemistry. By offering rigorously validated datasets, it enables systematic evaluation and refinement of emerging techniques, ultimately driving the progress in chemical information extraction.

\section{Acknowledgment}

This work supported by the Ministry of Economic Development of the Russian Federation (IGK 000000C313925P4C0002), agreement No139-15-2025-010.

We sincerely thank Olga Kononova for constructive feedback and fruitful discussions that helped us improve the manuscript.

{\small
\bibliographystyle{unsrt}
%\bibliography{references}

}

\appendix

\newpage
\section{ChemX ontology}
\label{ontology}

\begin{table}[h!]
\centering
\caption{ChemX benchmark datasets grouped by domain.}
\label{table_datasets}
\resizebox{\textwidth}{!}{%
\renewcommand{\arraystretch}{1.2}
\begin{tabular}{llllll}
\hline
\multicolumn{1}{c}{\multirow{2}{*}{\textbf{Domain}}} &
  \multicolumn{1}{c}{\multirow{2}{*}{\textbf{Dataset}}} &
  \multicolumn{1}{c}{\multirow{2}{*}{\textbf{Size}}} &
  \multicolumn{2}{c}{\textbf{Features}} &
  \multicolumn{1}{c}{\multirow{2}{*}{\textbf{Description}}} \\
\multicolumn{1}{c}{} &
  \multicolumn{1}{c}{} &
  \multicolumn{1}{c}{} &
  \multicolumn{1}{c}{\textbf{String}} &
  \multicolumn{1}{c}{\textbf{Numeric}} &
  \multicolumn{1}{c}{} \\ \hline 
 &
  Cytotox &
  5535 &
  12 &
  9 &
  \begin{tabular}[c]{@{}l@{}}Cytotoxicity of nanoparticles\\ in normal and cancer cell lines.\end{tabular} \\
 &
  Seltox &
  3286 &
  9 &
  14 &
  \begin{tabular}[c]{@{}l@{}}Toxic effects of nanoparticles\\ on bacterial strains.\end{tabular} \\
\begin{tabular}[c]{@{}l@{}}Nano-\\ materials\end{tabular} &
  Synergy &
  3326 &
  10 &
  19 &
  \begin{tabular}[c]{@{}l@{}}Drug–nanoparticle synergy\\ in antibacterial assays.\end{tabular} \\
 &
  Nanozymes &
  1135 &
  9 &
  11 &
  \begin{tabular}[c]{@{}l@{}}Catalytic properties of inorganic\\  enzyme mimics.\end{tabular} \\
 &
  Nanomag &
  2578 &
  8 &
  16 &
  \begin{tabular}[c]{@{}l@{}}Magnetic nanomaterials\\ and their biomedical uses.\end{tabular} \\ \hline
 &
  Benzimidazoles &
  1721 &
  6 &
  1 &
  \begin{tabular}[c]{@{}l@{}}SMILES molecules with MICs \\ for antibiotic SAR studies.\end{tabular} \\
 &
  Oxazolidinones &
  2923 &
  6 &
  1 &
  \begin{tabular}[c]{@{}l@{}}Synthetic antibiotics with \\ biological activity data.\end{tabular} \\ 
\begin{tabular}[c]{@{}l@{}}Small\\ molecules\end{tabular} &
  Complexes &
  907 &
  4 &
  1 &
  \begin{tabular}[c]{@{}l@{}} Organometallic chelate complexes \\ with thermodynamic parameters.\end{tabular} \\
 &
  Eye Drops &
  163 &
  2 &
  1 &
  \begin{tabular}[c]{@{}l@{}}Drug permeability data across \\ corneal tissue.\end{tabular} \\
 &
  Co-crystals &
  70 &
  7 &
  0 &
  \begin{tabular}[c]{@{}l@{}}Drug co-crystals with improved \\ photostability.\end{tabular} \\ \hline
\end{tabular}%
}
\end{table}

For small molecule datasets, the ontology centers around molecular descriptors, including SMILES representations, biological activity metrics (e.g., MIC, IC$_{50}$), and compound-specific metadata. In contrast, nanomaterials and other material-centric datasets involve a substantially broader set of parameters, encompassing physicochemical properties (e.g., size, zeta potential, surface coating), synthesis conditions, structural characteristics, and application-specific outcomes. This reflects the inherent complexity and multimodality of material-related information in scientific literature.

\subsection{Labeling datasets by complexity level for extraction}

\label{Selection of articles for analysis}

\begin{table}[h!]
\centering
\caption{Selection of articles for analysis. }
\label{table_atricles}
\renewcommand{\arraystretch}{1.3}
\begin{tabular}{lll}
\hline
Domain                           & \multicolumn{1}{c}{Dataset} & \multicolumn{1}{c}{Complexity} \\ \hline
\multirow{5}{*}{Nanomaterials}   & Cytotoxicity                & High                           \\
                                 & Seltox                      & High                           \\
                                 & Synergy                     & High                           \\
                                 & \textbf{Nanozymes}          & Medium                         \\
                                 & Nanomag                     & High                           \\ \hline
\multirow{5}{*}{Small molecules} & Benzimidazoles              & Medium                         \\
                                 & Oxazolidinones              & Medium                         \\
                                 & \textbf{Complexes}          & Low                            \\
                                 & Eye drops                   & Low                            \\
                                 & Co-crystalls                & Medium                         \\ \hline
\end{tabular}
\end{table}
We assess dataset extraction complexity with five interrelated criteria that capture common challenges in automated scientific data extraction. Heterogeneous information formats—continuous text, tables, and figures that often disperse related data and encode values in complex plots or schematics—make parsing difficult \citep{polak2024}. Non‑uniform table structures and cases where essential details appear only in the main text require cross‑referencing, while semantic ambiguity in parameter labels and variable units demands contextual inference for correct mapping \citep{polak2024}. Records with single numeric values are easier to extract reliably, whereas multi‑value records need careful linking of each value to the proper material and unit, increasing error risk. Finally, domain differences matter: inorganic nanomaterials frequently require hierarchical relationship extraction (composition and morphology → property), which is harder than extracting properties for small molecules that often use standardized encodings like SMILES \citep{hira2024}. 

Datasets are classified as low, medium, or high complexity based on these factors, with multi-format parsing, irregular tables, multi-value linking, and hierarchical relationships elevating difficulty (\autoref{table_atricles}).

\section{Quality Control}
\label{Quality Control}
\begin{figure}[h!]
    \centering
    \includegraphics[width=0.9\linewidth]{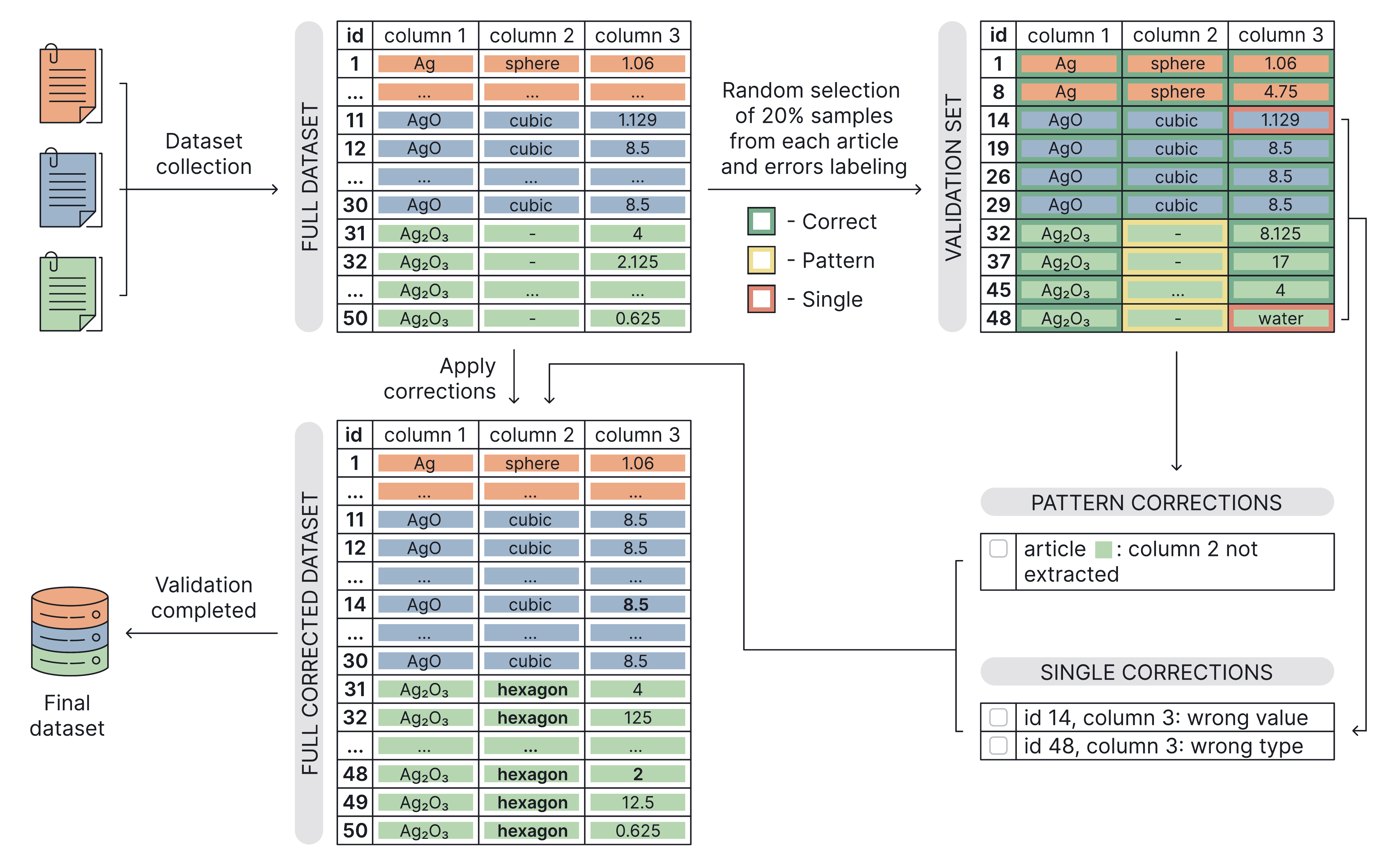}
    \caption{Quality control process for ChemX datasets}
    \label{fig:validation}
\end{figure}

A critical aspect of ChemX is its rigorous quality control process (\autoref{fig:validation}). To evaluate data integrity, we applied a stratified manual cross-verification procedure depicted on \autoref{fig:validation}. 

From each source article represented in a dataset, approximately 20\% of entries were randomly selected and reviewed against the original source material, including PDFs, figures, and supplementary tables. Sampling was rounded up to ensure that at least one entry from each source article was manually reviewed during the verification process.

Errors — including transcription mistakes, structural mismatches, unit inconsistencies, and unsupported inferences — were categorized as either common (recurring patterns) or isolated (single occurrence). Importantly, if an isolated error was identified during review, we systematically checked all the other entries from the same source article, even if they were not part of the original sample. This additional step was intended to determine whether similar issues occurred in other records from the same publication. In many cases, this allowed us to detect recurring patterns that were not evident in the initial sample, enabling the expansion of our correction rules beyond the reviewed subset. As a result, even single-instance errors had the potential to lead to pattern-based corrections across the dataset.

Error categorization informed the correction strategy. For common errors, we formulated rule-based recommendations that specified the field affected, the observed scope of recurrence, and the appropriate method for correction, such as structural replacement, unit standardization, or removal of inferred content. Corrections were then applied across the whole group. All recommendations were documented in writing and communicated to the dataset curators for implementation across relevant records. Isolated issues were corrected individually.

\section{Datasets Overview and Analysis}
\label{Datasets Overview and Analysis}

\begin{figure}[h!]
    \centering
    \includegraphics[width=0.9\linewidth]{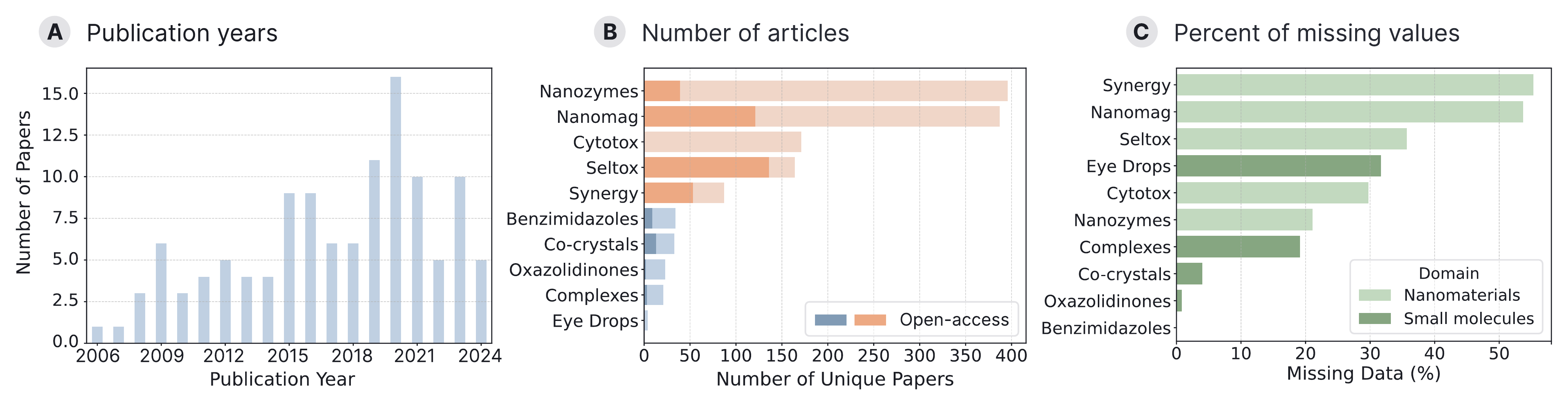}
    \caption{Quality control process for ChemX datasets}
    \label{fig:datasets}
\end{figure}

\autoref{fig:datasets}B shows the number of openly accessible articles per dataset. The publication year distribution (\autoref{fig:datasets}A) reflects literature growth since the early 2000s, with a sharp increase in the past decade. We also assessed missing values across datasets (\autoref{fig:datasets}C), with some exhibiting high sparsity due to incomplete reporting. This heterogeneity in data completeness benefits benchmarking by enabling rigorous evaluation of automated extraction systems—testing both accurate retrieval of reported values and correct identification of missing data.

\section{Experiments}

\subsection{Selected articles}
\label{selected_articles}

For each domain, we selected the two datasets of lowest complexity (nanozymes and complexes). For each dataset, three articles were picked for the experiments:

\begin{enumerate}
    \item Nanozymes
    \begin{enumerate}
        \item \href{https://doi.org/10.1155/2019/5416963}{Oxidase-Like Catalytic Performance of Nano-MnO2 and Its Potential Application for Metal Ions Detection in Water} \textbf{(Open Access)}
        \item \href{https://doi.org/10.1002/cbic.202000147}{Size Effect in Pd-Ir Core-Shell Nanoparticles as Nanozymes}
        \item \href{https://doi.org/10.1021/jacs.5b09337}{Single Nanoparticle to 3D Supercage: Framing for an Artificial Enzyme System}
    \end{enumerate}
    
    \item Complexes
    \begin{enumerate}
        \item \href{https://doi.org/10.1016/j.ccr.2022.214606}{Prediction of Gd(III) complex thermodynamic stability}
        \item \href{https://doi.org/10.1021/cr900325h}{Coordinating Radiometals of Copper, Gallium, Indium, Yttrium, and Zirconium for PET and SPECT Imaging of Disease}
        \item \href{https://doi.org/10.1590/S0103-50532006000800004}{Technetium and rhenium: coordination chemistry and nuclear medical applications} \textbf{(Open Access)}
    \end{enumerate}
\end{enumerate}
\subsection{Prompts and metrics}
\label{metrics}

For evaluating data extraction quality, we calculated the following:
\begin{itemize}
    \item \textbf{True Positives (TP):} The count of values correctly extracted (i.e., the value exists in both the original dataset and the extracted dataset).
    \item \textbf{False Positives (FP):} The count of values incorrectly extracted (i.e., the value does not exist in the original dataset but is present in the extracted dataset).
    \item \textbf{False Negatives (FN):} The count of missing values (i.e., the value exists in the original dataset but is absent from the extracted dataset).
\end{itemize}

For each PDF in the dataset, we computed precision, recall, and F1 score based on those quantities. The resulting metrics were then aggregated across all PDFs in the dataset and averaged by dividing the total sum by the number of PDFs.

To standardize inputs, we created the following prompt template:

\textbf{system\_prompt} = "You are a domain-specific chemical information extraction assistant. You specialize in the chemistry of ... . Your area of expertise includes ... ."

\textbf{user\_prompt} = "Your task is to extract **every** mention of ... for ... from a scientific article, and output a **JSON array** of objects **only** (no markdown, no commentary, no extra text):

\begin{enumerate}
    \item \textit{Feature 1 (string)}: \textit{Description} (e.g., \textit{'example'}).
    \item \textit{Feature 2 (numeric)}: \textit{Description} (e.g., \textit{'example'}).
    \item ...
    \item \textit{Target value (numeric)}: \textit{Description} (e.g., \textit{'example'}).
\end{enumerate}

Extraction rules:
\begin{itemize}
    \item Extract **each** ... mention as a separate object. 
    \item Do **not** filter, group, summarize, or deduplicate. Include repeated mentions and duplicates if they occur in different contexts. 
    \item If you cannot find a required field for an object, re-check the context; if it’s still absent, set that field’s value to {"NOT\_DETECTED"}
    \item \textit{Other rules specific to this dataset}
    \item The example of JSON below shows only one extracted samples, however your output should contain **all** mentions of ... for ... present in the article.
\end{itemize}

Output **must** be a single JSON array, like:
[\{ "feature 1": "example of feature 1", "feature
2": "example of feature 2", ... "target value": "example of target value" \}]"

\textbf{Complexes}

\textbf{system\_prompt} = "You are a domain-specific chemical information extraction assistant. You specialize in the chemistry of organometallic complexes and their properties."

\textbf{user\_prompt} = "Your task is to extract \textbf{every} mention of organometallic complexes and chelate ligands from scientific article, and output a \textbf{JSON array} of objects \textbf{only} (no markdown, no commentary, no extra text).

Fields for each object:
\begin{itemize}
    \item `compound\_id` (string): ID of a complex within the article, as cited in the text, e.g. `"L3"`, `"A31"`. 
    \item `compound\_name` (string): abbreviated or full name of the complex or ligand as cited in the text, e.g. `"DOTA"`, `"tebroxime"`. 
    \item `SMILES` (string): full SMILES representation of ligand environment or single ligand.  If a complete organometallic complex is shown, extract all ligand structures without mentioning the metal (e.g., "COc1cc(C=CC([O-])CC([O-])CC([O-])C=Cc2ccc(O)c(OC)c2)ccc1O. [C-]\#[O+].[C-]\#[O+].[C-]\#[O+].[OH-]"). For a chelate ligand without a complete organometallic complex, extract only that ligand's structure (e.g., 'O=C(O)CN(CCN(CC(CC(=O)O)CC(=O)O)CCN(CC(=O)O)CC(=O)O').
    \item `SMILES\_type` (string): one of `"ligand"` or `"environment"`. "environment" refers to the entire organometallic complex, including one or more ligands and a metal atom. 
    \item `target\_value` (number): the numeric value of logarithms of thermodynamic stability constants lgK or logK (without quotes). 
\end{itemize}

Extraction rules:
\begin{enumerate}
    \item Extract \textbf{each} mention of `target\_value` (lgK or logK) as a separate object. 
    \item Do \textbf{not} filter, group, summarize, or deduplicate. Include repeated mentions and duplicates if they occur in different contexts.  
    \item If a molecule is fully depicted in a figure, write it as a SMILES string. If a molecule is depicted as a scaffold and residues separately in different places of an article, connect them by compound ID or name into one molecule and write it a single SMILES string.
    \item If multiple thermodynamic stability constants appear for the same complex or ligand extract each separately. 
    \item  Extract only structures that comply with these rules:
        \begin{itemize}
            \item The complexes must contain \textbf{Ga} as the metal or the ligands must belong to complexes of that metal.
            \item The complete molecular structure shall be given without errors in it or identifiers.
            \item Compounds must contain more than one carbon (exclude CO, Me).
            \item Compounds must not contain polymeric structures, attached biomolecules or carboranes, undefined radicals, undeciphered designations (e.g., amino acids) beyond the simplest abbreviations (i.e., Me, Et, Pr, Bu, Ph, Ac), names of radicals instead of their structure, or incomplete indication of the ligand structure (e.g., L = P, N).
            \item Compounds must not be reaction intermediate or precursor.
        \end{itemize}
    \item If you cannot find a required field for an object, re-check the context; if it’s still absent, set that field’s value to `"NOT\_DETECTED"`.  
    \item The example of JSON below shows only two extracted samples, however your output should contain \textbf{all} mentions of organometallic complexes and / or chelate ligands present in the article.
\end{enumerate}

Output \textbf{must} be a single JSON array, like:
\newline [
    \{
        \newline "compound\_id": "L3",
        \newline "compound\_name": "DOTA",
        \newline "SMILES": "O=C(O)CN(CCN(CC(=O)O)CC(=O)O)CC(=O)O",
        \newline "SMILES\_type": "ligand",
        \newline "target": 21.3
    \},
    \{
        \newline "compound\_id": "A31",
        \newline "compound\_name": "tebroxime",
        \newline "SMILES": "[C-]\#[N+]CC(C)(C)OC.[C-]\#[N+]CC(C)(C)OC.[C-]\#[N+]CC(C)(C)OC.[C-]\#[N+]CC(C)(C)OC.[C-]\#[N+]CC(C)(C)OC.[C-]\#[N+]CC(C)(C)OC",
        \newline "SMILES\_type": "environment",
        \newline "target": 17.9
    \}
]"

\textbf{Nanozymes}

\textbf{system\_prompt} = "You are a domain-specific chemical information extraction assistant. You specialize in nanozymes."

\textbf{user\_prompt} = "Your task is to extract \textbf{every} mention of experiments for \textbf{ALL} nanozymes from a scientific article and output a \textbf{JSON array} of objects \textbf{only} (no markdown, no commentary, no extra text).

Fields for each object:
\begin{itemize}
    \item `formula` (string): the chemical formula of the nanozyme, e.g. "Fe3O4", "CuO", etc.
    \item `activity` (string): catalytic activity type, typically "peroxidase", "oxidase", "catalase", "laccase", or other.
    \item `syngony` (string): the crystal unit of the nanozyme, e.g. "cubic", "hexagonal", "tetragonal", "monoclinic", "orthorhombic", "trigonal", "amorphous", "triclinic".
    \item `length` (number): the length of the nanozyme particle in nanometers.
    \item `width` (number): the width of the nanozyme particle in nanometers.
    \item `depth` (number): the depth of the nanozyme particle in nanometers.
    \item `surface` (string): the molecule on the surface of the nanozyme, e.g., "naked", "poly(ethylene oxide)", "poly(N-Vinylpyrrolidone)", "Tetrakis(4-carboxyphenyl)porphine", or other.
    \item `km\_value` (number): the Michaelis constant value for the nanozyme.
    \item `km\_unit` (string): the unit for the Michaelis constant, e.g., "mM", etc.
    \item `vmax\_value` (number): the molar maximum reaction rate value.
    \item `vmax\_unit` (string): the unit for the maximum reaction rate, e.g., "$\mu$mol/min", "mol/min", etc.
    \item `reaction\_type` (string): the reaction type involving the substrate and co-substrate, e.g., "TMB + H2O2", "H2O2 + TMB", "TMB", "ABTS + H2O2", "H2O2", "OPD + H2O2", "H2O2 + GSH", or other.
    \item `c\_min` (number): the minimum substrate concentration in catalytic assays in mM.
    \item `c\_max` (number): the maximum substrate concentration in catalytic assays in mM.
    \item `c\_const` (number): the constant co-substrate concentration used during assays.
    \item `c\_const\_unit` (string): the unit of measurement for co-substrate concentration.
    \item `ccat\_value` (number): the concentration of the catalyst used in assays.
    \item `ccat\_unit` (string): the unit of measurement for catalyst concentration.
    \item `ph` (number): the pH level at which experiments were conducted.
    \item `temperature` (number): the temperature in Celsius during the study.
\end{itemize}

Extraction rules:
\begin{enumerate}
    \item Extract \textbf{each} nanozyme mention as a separate object. 
    \item Do \textbf{not} filter, group, summarize, or deduplicate. Include repeated mentions and duplicates if they occur in different contexts. 
    \item If you cannot find a required field for an object, re-check the context; if it’s still absent, set that field’s value to `"NOT\_DETECTED"`.
    \item The example of JSON below shows only two extracted samples, however your output should contain \textbf{all} nanozymes present in the article.
\end{enumerate}

Output \textbf{must} be a single JSON array, like:
\newline [
  \{
    \newline "formula": "Fe3O4",
    \newline "activity": "peroxidase",
    \newline "syngony": "cubic",
    \newline "length": 10,
    \newline "width": 10,
    \newline "depth": 2.5,
    \newline "surface": "naked",
    \newline "km\_value": 0.2,
    \newline "km\_unit": "mM",
    \newline "vmax\_value": 2.5,
    \newline "vmax\_unit": "µmol/min",
    \newline "reaction\_type": "TMB + H2O2",
    \newline "c\_min": 0.01,
    \newline "c\_max": 1.0,
    \newline "c\_const": 1.0,
    \newline "c\_const\_unit": "mM",
    \newline "ccat\_value": 0.05,
    \newline "ccat\_unit": "mg/mL",
    \newline "ph": 4.0,
    \newline "temperature": 25
  \},
  \{
    \newline "formula": "CeO2",
    \newline "activity": "oxidase",
    \newline "syngony": "cubic",
    \newline "length": 5,
    \newline "width": 5,
    \newline "depth": 200,
    \newline "surface": "poly(ethylene oxide)",
    \newline "km\_value": 54.05,
    \newline "km\_unit": "mM",
    \newline "vmax\_value": 7.88,
    \newline "vmax\_unit": "10-8 M s-1",
    \newline "reaction\_type": "TMB",
    \newline "c\_min": 0.02,
    \newline "c\_max": 0.8,
    \newline "c\_const": 800,
    \newline "c\_const\_unit": "$\mu$M",
    \newline "ccat\_value": 0.02,
    \newline "ccat\_unit": "mg/mL",
    \newline "ph": 5.5,
    \newline "temperature": 37
  \}
]"

\section{Results and Discussion}
\label{Results and Discussion}

% \usepackage{multirow}
\begin{table}[ht!]
\centering
\caption{All metrics for complexes dataset (baseline models).}
\label{tab:all_comp_baseline}
\renewcommand{\arraystretch}{1.2}
\begin{tabular}{lcccccc}
\hline
\multicolumn{1}{c}{\multirow{2}{*}{Column}} & \multicolumn{3}{c}{GPT-5} & \multicolumn{3}{c}{GPT-5 Thinking} \\ \cline{2-7} 
\multicolumn{1}{c}{}                        & Precision  & Recall  & F1 & Precision     & Recall     & F1    \\ \hline
compound\_id   & 0.65 & 0.29 & 0.35 & 0.65 & 0.52 & 0.58 \\
compound\_name & 0.41 & 0.22 & 0.26 & 0.44 & 0.37 & 0.40  \\
SMILES         & 0.14 & 0.03 & 0.04 & 0.00  & 0.00  & 0.00  \\
SMILES\_type   & 0.67 & 0.3  & 0.36 & 0.00  & 0.00  & 0.00  \\
target         & 0.41 & 0.1  & 0.14 & 0.00  & 0.00  & 0.00  \\ \hline
\end{tabular}
\end{table}

\begin{table}[ht!]
\centering
\caption{All metrics for complexes dataset (single-agent approach).}
\label{tab:all_comp_single}
\resizebox{\textwidth}{!}{%
\renewcommand{\arraystretch}{1.2}
\begin{tabular}{lccccccccc}
\hline
\multicolumn{1}{c}{\multirow{2}{*}{Column}} & \multicolumn{3}{c}{GPT-4.1} & \multicolumn{3}{c}{GPT-5} & \multicolumn{3}{c}{GPT-OSS-20b} \\ \cline{2-10} 
\multicolumn{1}{c}{}                        & Precision   & Recall  & F1  & Precision  & Recall  & F1 & Precision    & Recall    & F1   \\ \hline
compound\_id   & 0.56 & 0.35 & 0.43 & 0.73 & 0.88 & 0.80  & 0.74 & 0.63 & 0.68 \\
compound\_name & 0.13 & 0.08 & 0.1  & 0.05 & 0.06 & 0.05 & 0.07 & 0.06 & 0.07 \\
SMILES         & 0.06 & 0.04 & 0.05 & 0.00  & 0.00  & 0.00  & 0.00  & 0.00  & 0.00  \\
SMILES\_type   & 1.00  & 0.63 & 0.77 & 0.83 & 1.00  & 0.91 & 1.00  & 0.84 & 0.91 \\
target         & 0.00  & 0.00  & 0.00  & 0.00  & 0.00  & 0.00  & 0.00  & 0.00  & 0.00  \\ \hline
\end{tabular}%
}
\end{table}

\begin{table}[ht!]
\centering
\caption{All metrics for complexes dataset (multi-agent approaches).}
\label{tab:all_comp_multi}
\resizebox{\textwidth}{!}{%
\renewcommand{\arraystretch}{1.2}
\begin{tabular}{lccccccccc}
\hline
\multicolumn{1}{c}{\multirow{2}{*}{Column}} & \multicolumn{3}{c}{ChatGPT Agent} & \multicolumn{3}{c}{SLM Matrix} & \multicolumn{3}{c}{FutureHouse} \\ \cline{2-10} 
\multicolumn{1}{c}{}                        & Precision     & Recall    & F1    & Precision    & Recall   & F1   & Precision    & Recall    & F1   \\ \hline
compound\_id   & 0.64 & 0.52 & 0.57 & 0.93 & 0.89 & 0.91 & 0.06 & 0.06 & 0.06 \\
compound\_name & 0.50  & 0.41 & 0.45 & 0.11 & 0.11 & 0.11 & 0.00  & 0.00  & 0.00  \\
SMILES         & 0.06 & 0.04 & 0.05 & 0.01 & 0.01 & 0.01 & 0.00  & 0.00  & 0.00  \\
SMILES\_type   & 0.56 & 0.47 & 0.51 & 0.94 & 0.90  & 0.92 & 0.52 & 0.24 & 0.25 \\
target         & 0.73 & 0.67 & 0.70  & 0.00  & 0.00  & 0.00  & 0.04 & 0.00  & 0.00  \\ \hline
\end{tabular}%
}
\end{table}

% \usepackage{multirow}
\begin{table}[ht!]
\centering
\caption{All metrics for nanozymes dataset (baseline models).}
\label{tab:all_metrics_baseline}
\renewcommand{\arraystretch}{1.2}
\begin{tabular}{lllllll}
\hline
\multicolumn{1}{c}{\multirow{2}{*}{Column}} &
  \multicolumn{3}{c}{GPT-5} &
  \multicolumn{3}{c}{GPT-5 Thinking} \\ \cline{2-7} 
\multicolumn{1}{c}{} &
  \multicolumn{1}{c}{Precision} &
  \multicolumn{1}{c}{Recall} &
  \multicolumn{1}{c}{F1} &
  \multicolumn{1}{c}{Precision} &
  \multicolumn{1}{c}{Recall} &
  \multicolumn{1}{c}{F1} \\ \hline
formula        & 0.62 & 1.00  & 0.71 & 0.02 & 0.08 & 0.03 \\
activity       & 0.62 & 1.00  & 0.71 & 0.02 & 0.08 & 0.03 \\
syngony        & 0.62 & 1.00  & 0.71 & 0.02 & 0.08 & 0.03 \\
length         & 0.36 & 0.42 & 0.38 & 0.02 & 0.04 & 0.03 \\
width          & 0.25 & 0.25 & 0.25 & 0.02 & 0.02 & 0.01 \\
depth          & 0.47 & 0.58 & 0.52 & 0.01 & 0.02 & 0.01 \\
surface        & 0.00  & 0.00  & 0.00  & 0.00    & 0.03 & 0.00    \\
km\_value      & 0.07 & 0.33 & 0.11 & 0.01 & 0.05 & 0.02 \\
km\_unit       & 0.07 & 0.33 & 0.11 & 0.01 & 0.05 & 0.02 \\
vmax\_value    & 0.40 & 0.67 & 0.44 & 0.01 & 0.05 & 0.02 \\
vmax\_unit     & 0.40 & 0.67 & 0.44 & 0.01 & 0.05 & 0.02 \\
reaction\_type & 0.44 & 0.50  & 0.47 & 0.02 & 0.04 & 0.02 \\
c\_min         & 0.07 & 0.33 & 0.11 & 0.00   & 0.03 & 0.00    \\
c\_max         & 0.07 & 0.33 & 0.11 & 0.00    & 0.03 & 0.00    \\
c\_const       & 0.33 & 0.33 & 0.33 & 0.00    & 0.00    & 0.00    \\
c\_const\_unit & 0.40 & 0.67 & 0.44 & 0.01 & 0.05 & 0.02 \\
ccat\_value    & 0.46 & 0.83 & 0.54 & 0.00    & 0.04 & 0.01 \\
ccat\_unit     & 0.33 & 0.33 & 0.33 & 0.01 & 0.03 & 0.02 \\
ph             & 0.62 & 1.00  & 0.71 & 0.02 & 0.08 & 0.03 \\
temperature    & 0.00  & 0.00  & 0.00  & 0.00    & 0.00    & 0.00    \\ \hline
\end{tabular}
\end{table}

% \usepackage{multirow}
% \usepackage{graphicx}
\begin{table}[ht!]
\centering
\caption{All metrics for nanozymes dataset (Single-agent approach).}
\label{tab:all_metrics_single}
\resizebox{\textwidth}{!}{%
\renewcommand{\arraystretch}{1.2}
\begin{tabular}{lccccccccc}
\hline
\multicolumn{1}{c}{\multirow{2}{*}{Column}} & \multicolumn{3}{c}{GPT-4.1} & \multicolumn{3}{c}{GPT-5} & \multicolumn{3}{c}{GPT-OSS-20b} \\ \cline{2-10} 
\multicolumn{1}{c}{}                        & Precision   & Recall  & F1  & Precision  & Recall  & F1 & Precision    & Recall    & F1   \\ \hline
formula        & 0.56 & 1.00 & 0.71 & 0.62 & 1.00  & 0.71 & 0.83 & 1.00 & 0.91 \\
activity       & 0.56 & 1.00 & 0.71 & 0.62 & 1.00  & 0.71 & 0.83 & 1.00 & 0.91 \\
syngony        & 0.56 & 1.00 & 0.71 & 0.62 & 1.00  & 0.71 & 0.17 & 0.20 & 0.18 \\
length         & 0.44 & 0.80 & 0.57 & 0.36 & 0.42 & 0.38 & 0.67 & 0.80 & 0.73 \\
width          & 0.11 & 0.20 & 0.14 & 0.25 & 0.25 & 0.25 & 0.67 & 0.80 & 0.73 \\
depth          & 0.11 & 0.20 & 0.14 & 0.47 & 0.58 & 0.52 & 0.67 & 0.80 & 0.73 \\
surface        & 0.00  & 0.00 & 0.00  & 0.00  & 0.00  & 0.00  & 0.00  & 0.00 & 0.00  \\
km\_value      & 0.56 & 1.00 & 0.71 & 0.07 & 0.33 & 0.11 & 0.83 & 1.00 & 0.91 \\
km\_unit       & 0.44 & 0.80 & 0.57 & 0.07 & 0.33 & 0.11 & 0.67 & 0.80 & 0.73 \\
vmax\_value    & 0.56 & 1.00 & 0.71 & 0.40  & 0.67 & 0.44 & 0.83 & 1.00 & 0.91 \\
vmax\_unit     & 0.44 & 0.80 & 0.57 & 0.40  & 0.67 & 0.44 & 0.67 & 0.80 & 0.73 \\
reaction\_type & 0.56 & 1.00 & 0.71 & 0.44 & 0.50  & 0.47 & 0.67 & 0.80 & 0.73 \\
c\_min         & 0.44 & 0.80 & 0.57 & 0.07 & 0.33 & 0.11 & 0.17 & 0.20 & 0.18 \\
c\_max         & 0.44 & 0.80 & 0.57 & 0.07 & 0.33 & 0.11 & 0.17 & 0.20 & 0.18 \\
c\_const       & 0.44 & 0.80 & 0.57 & 0.33 & 0.33 & 0.33 & 0.67 & 0.80 & 0.73 \\
c\_const\_unit & 0.56 & 1.00 & 0.71 & 0.40  & 0.67 & 0.44 & 0.67 & 0.80 & 0.73 \\
ccat\_value    & 0.33 & 0.60 & 0.43 & 0.46 & 0.83 & 0.54 & 0.50  & 0.60 & 0.55 \\
ccat\_unit     & 0.44 & 0.80 & 0.57 & 0.33 & 0.33 & 0.33 & 0.67 & 0.80 & 0.73 \\
ph             & 0.56 & 1.00 & 0.71 & 0.62 & 1.00  & 0.71 & 0.83 & 1.00 & 0.91 \\
temperature    & 0.00  & 0.00 & 0.00  & 0.00  & 0.00  & 0.00  & 0.00  & 0.00 & 0.00  \\ \hline
\end{tabular}%
}
\end{table}

% \usepackage{multirow}
% \usepackage{graphicx}
\begin{table}[ht!]
\centering
\caption{All metrics for nanozymes dataset (Multi-agent approaches).}
\label{tab:all_metrics_multi}
\resizebox{\textwidth}{!}{%
\renewcommand{\arraystretch}{1.2}
\begin{tabular}{lccccccccc}
\hline
\multicolumn{1}{c}{\multirow{2}{*}{Column}} & \multicolumn{3}{c}{SLM-Matrix} & \multicolumn{3}{c}{FutureHouse} & \multicolumn{3}{c}{NanoMINER} \\ \cline{2-10} 
\multicolumn{1}{c}{}                        & Precision    & Recall   & F1   & Precision    & Recall    & F1   & Precision   & Recall   & F1   \\ \hline
formula        & 0.25 & 1.00 & 0.40  & 0.12 & 0.67 & 0.21 & -    & -    & -    \\
activity       & 0.25 & 1.00 & 0.40  & 0.12 & 0.67 & 0.21 & -    & -    & -    \\
syngony        & 0.05 & 0.20 & 0.08 & 0.08 & 0.50  & 0.14 & -    & -    & -    \\
length         & 0.20  & 0.80 & 0.32 & 0.04 & 0.17 & 0.07 & -    & -    & -    \\
width          & 0.20  & 0.80 & 0.32 & 0.00  & 0.00  & 0.00  & -    & -    & -    \\
depth          & 0.20  & 0.80 & 0.32 & 0.04 & 0.17 & 0.07 & -    & -    & -    \\
surface        & 0.20  & 0.80 & 0.32 & 0.08 & 0.33 & 0.13 & -    & -    & -    \\
km\_value      & 0.05 & 0.20 & 0.08 & 0.04 & 0.33 & 0.07 & 0.97 & 0.91 & 0.94 \\
km\_unit       & 0.05 & 0.20 & 0.08 & 0.08 & 0.50  & 0.14 & -    & -    & -    \\
vmax\_value    & 0.05 & 0.20 & 0.08 & 0.04 & 0.33 & 0.07 & 0.96 & 0.83 & 0.89 \\
vmax\_unit     & 0.05 & 0.20 & 0.08 & 0.04 & 0.33 & 0.07 & -    & -    & -    \\
reaction\_type & 0.20  & 0.80 & 0.32 & 0.04 & 0.17 & 0.07 & -    & -    & -    \\
c\_min         & 0.05 & 0.20 & 0.08 & 0.04 & 0.33 & 0.07 & 0.97 & 0.54 & 0.69 \\
c\_max         & 0.05 & 0.20 & 0.08 & 0.08 & 0.50  & 0.14 & 0.97 & 0.53 & 0.69 \\
c\_const       & 0.20  & 0.80 & 0.32 & 0.00  & 0.00  & 0.00  & 0.78 & 0.51 & 0.62 \\
c\_const\_unit & 0.20  & 0.80 & 0.32 & 0.04 & 0.33 & 0.07 & -    & -    & -    \\
ccat\_value    & 0.05 & 0.20 & 0.08 & 0.04 & 0.33 & 0.07 & 0.88 & 0.81 & 0.84 \\
ccat\_unit     & 0.00  & 0.00 & 0.00  & 0.00  & 0.00  & 0.00  & -    & -    & -    \\
ph             & 0.25 & 1.00 & 0.40  & 0.12 & 0.67 & 0.21 & 0.98 & 0.82 & 0.89 \\
temperature    & 0.20  & 0.80 & 0.32 & 0.00  & 0.00  & 0.00  & 0.70  & 0.96 & 0.81 \\ \hline
\end{tabular}%
}
\end{table}

\section{Limitations}

While this benchmark encompasses ten datasets across two chemical domains, its scope is necessarily constrained and does not extend to other critical areas of chemistry, including organic reaction schemes, spectral data, quantum chemical calculations, and others.

Our experimental results on structure extraction underscore the inherent limitations of both general-purpose large language models (LLMs) and agent-based methodologies for the specific task of chemical structure recognition. Furthermore, even specialized agent-based systems demonstrated suboptimal performance. Although dedicated tools such as DECIMER \citep{Rajan2023} exist for converting molecular images into SMILES strings, their practical integration into automated extraction pipelines is presently precluded by two unresolved technical challenges: (1) the reliable detection of individual molecular images within the complex layouts of scientific articles, and (2) the accurate segmentation of images exhibiting heterogeneous formats and styles. Future advancements in computer vision, particularly in automated molecular localization and standardized image preprocessing, may eventually facilitate the incorporation of such tools. However, due to these extant limitations, tools like DECIMER were deliberately excluded from the present experimental framework. It is critical to note that the incorrect extraction of chemical structures poses significant risks; hallucinations or errors can propagate through automated workflows, leading to failures in reproducibility, invalid computational results, and ultimately, the generation of erroneous scientific data.

%%%%%%%%%%%%%%%%%%%%%%%%%%%%%%%%%%%%%%%%%%%%%%%%%%%%%%%%%%%%

\newpage
\section*{NeurIPS Paper Checklist}

\begin{enumerate}
\item {\bf Claims}
    \item[] Question: Do the main claims made in the abstract and introduction accurately reflect the paper's contributions and scope?
    \item[] Answer: \answerYes{} % Replace by \answerYes{}, \answerNo{}, or \answerNA{}.
    \item[] Justification: The abstract and introduction clearly describe the release of ChemX, a curated benchmark of 10 datasets for automated information extraction in chemistry, and the evaluation of both mono- and multi-agent LLM-based systems.

\item {\bf Limitations}
    \item[] Question: Does the paper discuss the limitations of the work performed by the authors?
    \item[] Answer: \answerYes{} % Replace by \answerYes{}, \answerNo{}, or \answerNA{}.
    \item[] Justification: Section F discusses multiple limitations.

\item {\bf Theory assumptions and proofs}
    \item[] Question: For each theoretical result, does the paper provide the full set of assumptions and a complete (and correct) proof?
    \item[] Answer: \answerNA{} % Replace by \answerYes{}, \answerNo{}, or \answerNA{}.
    \item[] Justification: The paper does not include theoretical results.

    \item {\bf Experimental result reproducibility}
    \item[] Question: Does the paper fully disclose all the information needed to reproduce the main experimental results of the paper to the extent that it affects the main claims and/or conclusions of the paper (regardless of whether the code and data are provided or not)?
    \item[] Answer: \answerYes{} % Replace by \answerYes{}, \answerNo{}, or \answerNA{}.
    \item[] Justification: In this article, we provide full documentation for each dataset, describe the methodology of the extraction experiments, and also include the code for conducting these experiments in Sections 4.

\item {\bf Open access to data and code}
    \item[] Question: Does the paper provide open access to the data and code, with sufficient instructions to faithfully reproduce the main experimental results, as described in the supplemental material?
    \item[] Answer: \answerYes{} % Replace by \answerYes{}, \answerNo{}, or \answerNA{}.
    \item[] Justification: Datasets and code are available via HuggingFace and GitHub with accompanying documentation.

\item {\bf Experimental setting/details}
    \item[] Question: Does the paper specify all the training and test details (e.g., data splits, hyperparameters, how they were chosen, type of optimizer, etc.) necessary to understand the results?
    \item[] Answer: \answerYes{} % Replace by \answerYes{}, \answerNo{}, or \answerNA{}.
    \item[] Justification: Sections 4 and D outline LLM setup, prompt structure, document formats, and evaluation procedures.

\item {\bf Experiment statistical significance}
    \item[] Question: Does the paper report error bars suitably and correctly defined or other appropriate information about the statistical significance of the experiments?
    \item[] Answer: \answerNo{} % Replace by \answerYes{}, \answerNo{}, or \answerNA{}.
    \item[] Justification: Experimental errors were not incorporated into the analysis, as the central claim of this work is not the comparative performance of the methods. Rather, we assert that all evaluated methods perform inadequately for the task. Consequently, the consideration of measurement error is immaterial, as its inclusion would not alter this overarching conclusion.

\item {\bf Experiments compute resources}
    \item[] Question: For each experiment, does the paper provide sufficient information on the computer resources (type of compute workers, memory, time of execution) needed to reproduce the experiments?
    \item[] Answer: \answerYes{} % Replace by \answerYes{}, \answerNo{}, or \answerNA{}.
    \item[] Justification: Experiments involving large language models such as GPT-4o were executed via the OpenAI API. All other computations, including preprocessing, single-agent pipeline execution, and evaluation metrics, were performed locally on a laptop with the following specifications: Intel Core i7-11800H (8 cores, 2.3–4.6 GHz), 16 GB RAM, and a 512 GB SSD. The GPU was not used for local execution.
    
\item {\bf Code of ethics}
    \item[] Question: Does the research conducted in the paper conform, in every respect, with the NeurIPS Code of Ethics \url{https://neurips.cc/public/EthicsGuidelines}?
    \item[] Answer: \answerYes{} % Replace by \answerYes{}, \answerNo{}, or \answerNA{}.
    \item[] Justification: All content was extracted from publicly accessible scientific literature or subscription-based academic access with proper institutional rights. No sensitive data or human participants were involved.

\item {\bf Broader impacts}
    \item[] Question: Does the paper discuss both potential positive societal impacts and negative societal impacts of the work performed?
    \item[] Answer: \answerYes{} % Replace by \answerYes{}, \answerNo{}, or \answerNA{}.
    \item[] Justification: Section F discusses the risks of incorrect extraction, hallucination in chemical contexts, and implications for reproducibility and automation in cheminformatics.
    
\item {\bf Safeguards}
    \item[] Question: Does the paper describe safeguards that have been put in place for responsible release of data or models that have a high risk for misuse (e.g., pretrained language models, image generators, or scraped datasets)?
    \item[] Answer: \answerNA{} % Replace by \answerYes{}, \answerNo{}, or \answerNA{}.
    \item[] Justification: No high-risk pretrained models or internet-scraped data were released.

\item {\bf Licenses for existing assets}
    \item[] Question: Are the creators or original owners of assets (e.g., code, data, models), used in the paper, properly credited and are the license and terms of use explicitly mentioned and properly respected?
    \item[] Answer: \answerYes{} % Replace by \answerYes{}, \answerNo{}, or \answerNA{}.
    \item[] Justification: The dataset is manually extracted from open-access and subscription-based articles accessed under institutional license, and all external tools and models are properly cited.

\item {\bf New assets}
    \item[] Question: Are new assets introduced in the paper well documented and is the documentation provided alongside the assets?
    \item[] Answer: \answerYes{} % Replace by \answerYes{}, \answerNo{}, or \answerNA{}.
    \item[] Justification: All 10 datasets are fully documented with schemas, annotation examples, and feature descriptions in the supplementary material and HuggingFace page.

\item {\bf Crowdsourcing and research with human subjects}
    \item[] Question: For crowdsourcing experiments and research with human subjects, does the paper include the full text of instructions given to participants and screenshots, if applicable, as well as details about compensation (if any)? 
    \item[] Answer: \answerNA{} % Replace by \answerYes{}, \answerNo{}, or \answerNA{}.
    \item[] Justification: No human participants or crowdworkers were involved in data collection or validation.

\item {\bf Institutional review board (IRB) approvals or equivalent for research with human subjects}
    \item[] Question: Does the paper describe potential risks incurred by study participants, whether such risks were disclosed to the subjects, and whether Institutional Review Board (IRB) approvals (or an equivalent approval/review based on the requirements of your country or institution) were obtained?
    \item[] Answer: \answerNA{} % Replace by \answerYes{}, \answerNo{}, or \answerNA{}.
    \item[] Justification: No human subjects were involved in the study. All data were derived from published scientific literature and manually curated by the authors.

\item {\bf Declaration of LLM usage}
    \item[] Question: Does the paper describe the usage of LLMs if it is an important, original, or non-standard component of the core methods in this research? Note that if the LLM is used only for writing, editing, or formatting purposes and does not impact the core methodology, scientific rigorousness, or originality of the research, declaration is not required.
    %this research? 
    \item[] Answer: \answerYes{} % Replace by \answerYes{}, \answerNo{}, or \answerNA{}.
    \item[] Justification: The paper explicitly discusses GPT-4.1, GPT-5, GPT-OSS-20b use for both baseline model and single-agent pipelines in Section 4.

\end{enumerate}

\end{document}